\documentclass{article} 
\usepackage{iclr2017_conference,times}

\usepackage{hyperref}
\usepackage{url}

\usepackage{color}
\usepackage{amsmath}
\usepackage{amsfonts}
\usepackage{nicefrac}

\usepackage{graphicx}
\usepackage{epstopdf}
\usepackage{subfigure}

\usepackage{tikz}
\usetikzlibrary{arrows}

\usepackage{xcolor}
\definecolor{darkgreen}{rgb}{0,0.6,0}
\definecolor{darkred}{rgb}{0.6,0,0}
\definecolor{darkblue}{rgb}{0,0.05,0.35}

\newcommand*\samethanks[1][\value{footnote}]{\footnotemark[#1]}
\newcommand{\ignore}[1]{}

\DeclareMathOperator*{\argmax}{arg\,max}

\title{Categorical Reparameterization\\ with Gumbel-Softmax}

\author{Eric Jang\\
Google Brain\\
\texttt{ejang@google.com} \\
\And
Shixiang Gu\thanks{Work done during an internship at Google Brain.}\\
University of Cambridge \\
MPI T\"{u}bingen \\
\texttt{sg717@cam.ac.uk} \\
\And
Ben Poole\samethanks\\ 
Stanford University \\
\texttt{poole@cs.stanford.edu}
}

\iclrfinalcopy 

\begin{document}
\maketitle

\begin{abstract}
Categorical variables are a natural choice for representing discrete structure in the world. However, stochastic neural networks rarely use categorical latent variables due to the inability to backpropagate through samples. In this work, we present an efficient gradient estimator that replaces the non-differentiable sample from a categorical distribution with a differentiable sample from a novel Gumbel-Softmax distribution. This distribution has the essential property that it can be smoothly annealed into a categorical distribution. We show that our Gumbel-Softmax estimator outperforms state-of-the-art gradient estimators on structured output prediction and unsupervised generative modeling tasks with categorical latent variables, and enables large speedups on semi-supervised classification.
\end{abstract}

\section{Introduction}

Stochastic neural networks with discrete random variables are a powerful technique for representing distributions encountered in unsupervised learning, language modeling, attention mechanisms, and reinforcement learning domains. For example, discrete variables have been used to learn probabilistic latent representations that correspond to distinct semantic classes \citep{kingma_ssvae}, image regions \citep{DBLP:journals/corr/XuBKCCSZB15}, and memory locations \citep{DBLP:journals/corr/GravesWD14,graves2016hybrid}. Discrete representations are often more interpretable \citep{DBLP:journals/corr/ChenDHSSA16} and more computationally efficient \citep{2016arXiv161009027R} than their continuous analogues.

However, stochastic networks with discrete variables are difficult to train because the backpropagation algorithm --- while permitting efficient computation of parameter gradients --- cannot be applied to non-differentiable layers. Prior work on stochastic gradient estimation has traditionally focused on either score function estimators augmented with Monte Carlo variance reduction techniques \citep{2012arXiv1206.6430P,NVIL,gu2016,DARN}, or biased path derivative estimators for Bernoulli variables \citep{DBLP:journals/corr/BengioLC13}. However, no existing gradient estimator has been formulated specifically for categorical variables. The contributions of this work are threefold:
\begin{enumerate}
  \item We introduce Gumbel-Softmax, a continuous distribution on the simplex that can approximate categorical samples, and whose parameter gradients can be easily computed via the reparameterization trick.
  \item We show experimentally that Gumbel-Softmax outperforms all single-sample gradient estimators on both Bernoulli variables and categorical variables.
  \item We show that this estimator can be used to efficiently train semi-supervised models (e.g. \citet{kingma_ssvae}) without costly marginalization over unobserved categorical latent variables. 
\end{enumerate}

The practical outcome of this paper is a simple, differentiable approximate sampling mechanism for categorical variables that can be integrated into neural networks and trained using standard backpropagation.

\section{The Gumbel-Softmax distribution}
We begin by defining the Gumbel-Softmax distribution, a continuous distribution over the simplex that can approximate samples from a categorical distribution. 
Let $z$ be a categorical variable with class probabilities $\pi_1,\pi_2,...\pi_k$. For the remainder of this paper we assume categorical samples are encoded as $k$-dimensional one-hot vectors lying on the corners of the $(k-1)$-dimensional simplex, $\Delta^{k-1}$. This allows us to define quantities such as the element-wise mean $\mathbb{E}_p[z] = \left[\pi_1, ..., \pi_k\right]$ of these vectors.

The Gumbel-Max trick \citep{13445,maddison2014sampling} provides a simple and efficient way to draw samples $z$ from a categorical distribution with class probabilities $\pi$: 
\begin{equation} \label{eq:gumbel}
z = \verb|one_hot|\left(\argmax_{i}{\left[ g_i + \log \pi_i \right]}\right)
\end{equation}
where $g_1...g_k$ are i.i.d samples drawn from $\text{Gumbel}(0,1)$\footnote{The $\text{Gumbel}(0,1)$ distribution can be sampled using inverse transform sampling by drawing $u \sim \text{Uniform}(0, 1)$ and computing $g= -\log(-\log(\text{u}))$. }.
We use the softmax function as a continuous, differentiable approximation to $\argmax$, and generate $k$-dimensional sample vectors $y \in \Delta^{k-1}$ where
\begin{equation}
y_i = \frac{\text{exp}((\log(\pi_i)+g_i)/\tau)}{\sum_{j=1}^k \text{exp}((\log(\pi_j)+g_j)/\tau)} \qquad \text{for } i=1, ..., k.
\end{equation}
The density of the Gumbel-Softmax distribution (derived in Appendix \ref{appendix:gumbel_derivation}) is:
\begin{equation}
p_{\pi, \tau}(y_1, ..., y_k) = \Gamma(k)\tau^{k-1}\left(\sum_{i=1}^{k} \pi_i/{y_i^\tau}\right)^{-k} \prod_{i=1}^k\left(\pi_i/y_i^{\tau+1}\right)\\
\end{equation}

This distribution was independently discovered by \citet{Maddison2016}, where it is referred to as the concrete distribution. As the softmax temperature $\tau$ approaches $0$, samples from the Gumbel-Softmax distribution become one-hot and the Gumbel-Softmax distribution becomes identical to the categorical distribution $p(z)$.

\begin{figure}[h]
  \centering
  \hspace{-5mm}
\includegraphics[width=.9\textwidth]{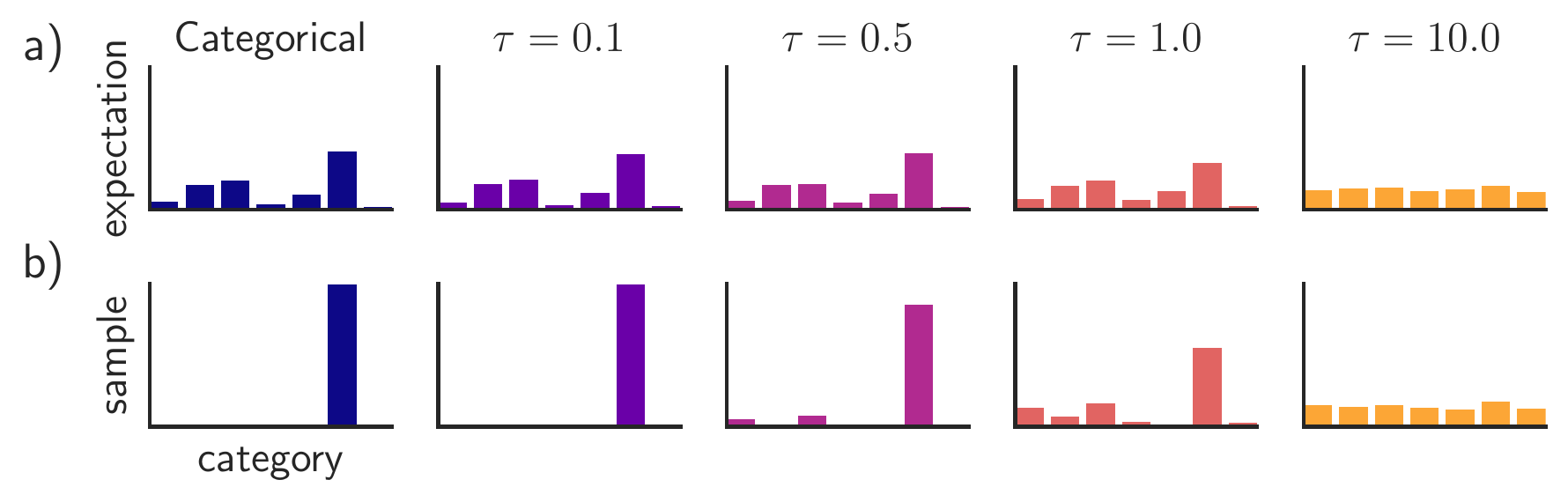}
  \caption{The Gumbel-Softmax distribution interpolates between discrete one-hot-encoded categorical distributions and continuous categorical densities. (a) For low temperatures ($\tau=0.1,\tau=0.5$), the expected value of a Gumbel-Softmax random variable approaches the expected value of a categorical random variable with the same logits. As the temperature increases ($\tau=1.0,\tau=10.0$), the expected value converges to a uniform distribution over the categories. (b) Samples from Gumbel-Softmax distributions are identical to samples from a categorical distribution as $\tau \to 0$. At higher temperatures, Gumbel-Softmax samples are no longer one-hot, and become uniform as $\tau \to \infty$. }
 \label{fig:gumbel_temp}
\end{figure}

\subsection{Gumbel-Softmax Estimator}

The Gumbel-Softmax distribution is smooth for $\tau > 0$, and therefore has a well-defined gradient $\nicefrac{\partial y}{\partial \pi}$ with respect to the parameters $\pi$. Thus, by replacing categorical samples with Gumbel-Softmax samples we can use backpropagation to compute gradients (see Section \ref{path_deriv}). We denote this procedure of replacing non-differentiable categorical samples with a differentiable approximation during training as the Gumbel-Softmax estimator.

While Gumbel-Softmax samples are differentiable, they are not identical to samples from the corresponding categorical distribution for non-zero temperature. 
For learning, there is a tradeoff between small temperatures, where samples are close to one-hot but the variance of the gradients is large, and large temperatures, where samples are smooth but the variance of the gradients is small (Figure \ref{fig:gumbel_temp}). In practice, we start at a high temperature and anneal to a small but non-zero temperature.

In our experiments, we find that the softmax temperature $\tau$ can be annealed according to a variety of schedules and still perform well. If $\tau$ is a learned parameter (rather than annealed via a fixed schedule), this scheme can be interpreted as entropy regularization \citep{szegedy2015rethinking,entropyreg}, where the Gumbel-Softmax distribution can adaptively adjust the ``confidence'' of proposed samples during the training process.

\subsection{Straight-Through Gumbel-Softmax Estimator}

Continuous relaxations of one-hot vectors are suitable for problems such as learning hidden representations and sequence modeling. For scenarios in which we are constrained to sampling discrete values (e.g. from a discrete action space for reinforcement learning, or quantized compression), we discretize $y$ using $\argmax$ but use our continuous approximation in the backward pass by approximating $\nabla_\theta z \approx \nabla_\theta y$. We call this the Straight-Through (ST) Gumbel Estimator, as it is reminiscent of the biased path derivative estimator described in \citet{DBLP:journals/corr/BengioLC13}. ST Gumbel-Softmax allows samples to be sparse even when the temperature $\tau$ is high.

\section{Related Work} \label{background}

\begin{figure}[h]  
  \centering
\includegraphics[width=\textwidth]{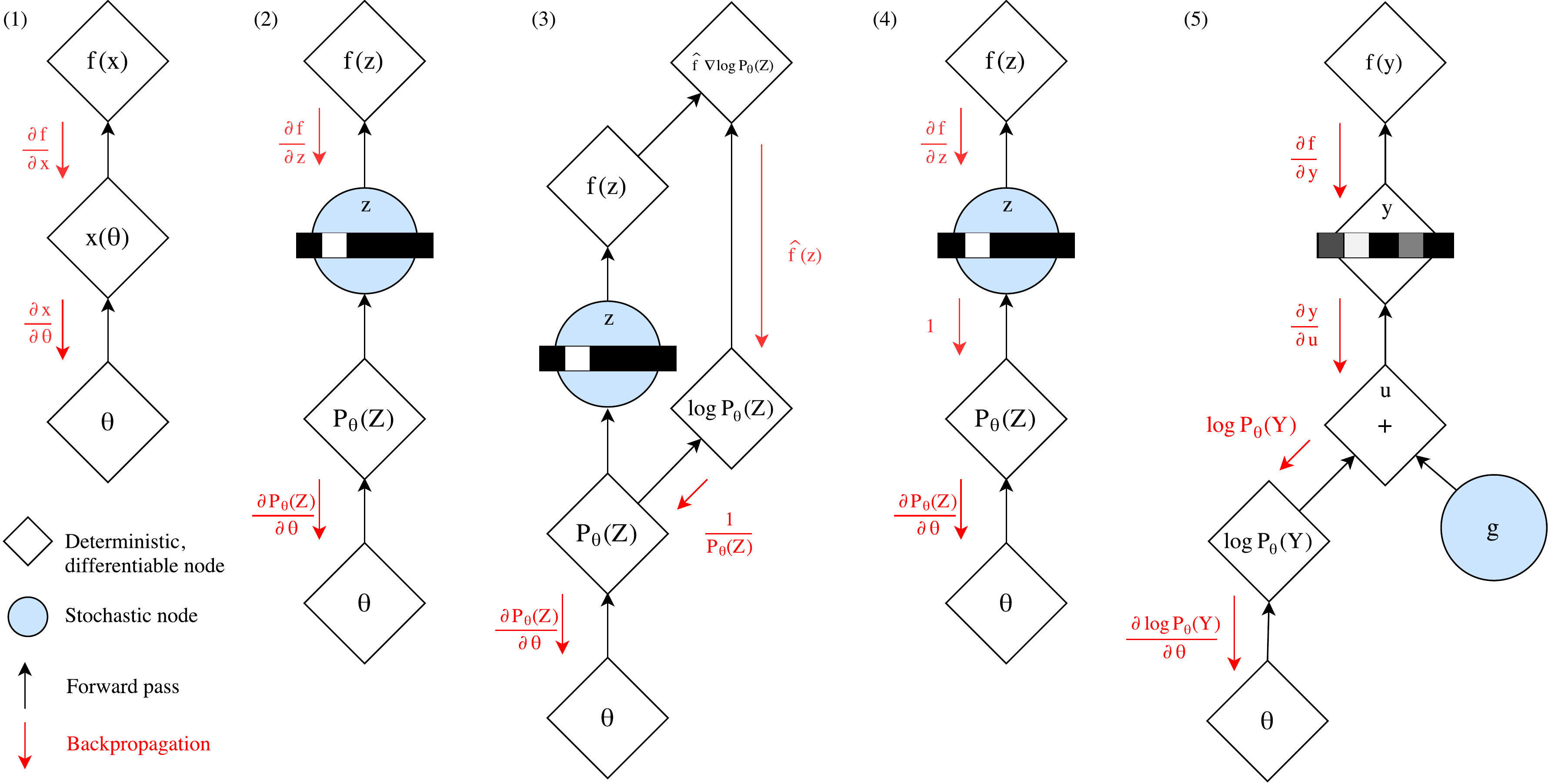}
  \caption{Gradient estimation in stochastic computation graphs. (1) $\nabla_\theta{f(x)}$ can be computed via backpropagation if $x(\theta)$ is deterministic and differentiable. (2) The presence of stochastic node $z$ precludes backpropagation as the sampler function does not have a well-defined gradient. (3) The score function estimator and its variants (NVIL, DARN, MuProp, VIMCO) obtain an unbiased estimate of $\nabla_\theta{f(x)}$ by backpropagating along a surrogate loss $\hat{f}\log p_\theta(z)$, where $\hat{f} = f(x) - b$ and $b$ is a baseline for variance reduction. (4) The Straight-Through estimator, developed primarily for Bernoulli variables, approximates $\nabla_\theta z \approx 1$. (5) Gumbel-Softmax is a path derivative estimator for a continuous distribution $y$ that approximates $z$. Reparameterization allows gradients to flow from $f(y)$ to $\theta$. $y$ can be annealed to one-hot categorical variables over the course of training.}
 \label{fig:stochastic_backprop}
\end{figure}

In this section we review existing stochastic gradient estimation techniques for discrete variables (illustrated in Figure \ref{fig:stochastic_backprop}). Consider a stochastic computation graph \citep{2015arXiv150605254S} with discrete random variable $z$ whose distribution depends on parameter $\theta$, and cost function $f(z)$. The objective is to minimize the expected cost $L(\theta) = \mathbb{E}_{z \sim p_\theta(z)}[f(z)]$ via gradient descent, which requires us to estimate $\nabla_\theta \mathbb{E}_{z \sim p_\theta(z)}[f(z)]$.

\subsection{Path Derivative Gradient Estimators}
\label{path_deriv}
For distributions that are reparameterizable, we can compute the sample $z$ as a deterministic function $g$ of the parameters $\theta$ and an independent random variable $\epsilon$, so that $z = g(\theta, \epsilon)$. The path-wise gradients from $f$ to $\theta$ can then be computed without encountering any stochastic nodes:
\begin{equation}
\frac{\partial}{\partial \theta} \mathbb{E}_{z\sim p_\theta}\left[f(z))\right] = \frac{\partial}{\partial \theta} \mathbb{E}_{\epsilon}\left[f(g(\theta,\epsilon))\right] = \mathbb{E}_{\epsilon\sim p_\epsilon}\left[\frac{\partial f}{\partial g} \frac{\partial g}{\partial \theta}\right]
\end{equation}

For example, the normal distribution $z \sim \mathcal{N}(\mu,\sigma)$ can be re-written as $\mu+\sigma \cdot \mathcal{N}(0,1)$, making it trivial to compute $\nicefrac{\partial z}{\partial \mu}$ and $\nicefrac{\partial z}{\partial \sigma}$. 
This reparameterization trick is commonly applied to training variational autooencoders with continuous latent variables using backpropagation \citep{kingma_vae,rezende2014}. As shown in Figure \ref{fig:stochastic_backprop}, we exploit such a trick in the construction of the Gumbel-Softmax estimator.

Biased path derivative estimators can be utilized even when $z$ is not reparameterizable. In general, we can approximate $\nabla_\theta z \approx \nabla_\theta m(\theta)$, where $m$ is a differentiable proxy for the stochastic sample. For Bernoulli variables with mean parameter $\theta$, the Straight-Through (ST) estimator \citep{DBLP:journals/corr/BengioLC13} approximates $m = \mu_\theta(z)$, implying $\nabla_\theta m = 1$. For $k=2$ (Bernoulli), ST Gumbel-Softmax is similar to the slope-annealed Straight-Through estimator proposed by \citet{HMRNN}, but uses a softmax instead of a hard sigmoid to determine the slope. \citet{Rolfe2016} considers an alternative approach where each binary latent variable parameterizes a continuous mixture model. Reparameterization gradients are obtained by backpropagating through the continuous variables and marginalizing out the binary variables.

One limitation of the ST estimator is that backpropagating with respect to the sample-independent mean may cause discrepancies between the forward and backward pass, leading to higher variance. Gumbel-Softmax avoids this problem because each sample $y$ is a differentiable proxy of the corresponding discrete sample $z$.

\subsection{Score Function-Based Gradient Estimators}

The score function estimator (SF, also referred to as REINFORCE \citep{Williams92simplestatistical} and likelihood ratio estimator \citep{LR}) uses the identity $\nabla_\theta p_\theta (z) = p_\theta (z) \nabla_\theta \log p_\theta (z)$ to derive the following unbiased estimator:

\begin{equation} \label{eq:sf}
\nabla_\theta \mathbb{E}_{z}\left[f(z)\right] = \mathbb{E}_z\left[f(z)\nabla_\theta \log p_\theta(z)\right]
\end{equation}

SF only requires that $p_\theta(z)$ is continuous in $\theta$, and does not require backpropagating through $f$ or the sample $z$. However, SF suffers from high variance and is consequently slow to converge. In particular, the variance of SF scales linearly with the number of dimensions of the sample vector \citep{2014arXiv1401.4082J}, making it especially challenging to use for categorical distributions.

The variance of a score function estimator can be reduced by subtracting a control variate $b(z)$ from the learning signal $f$, and adding back its analytical expectation $\mu_b = \mathbb{E}_z\left[b(z)\nabla_\theta \log p_\theta(z)\right]$ to keep the estimator unbiased:

\begin{align}
\nabla_\theta \mathbb{E}_{z}\left[f(z)\right] & = \mathbb{E}_z\left[f(z)\nabla_\theta \log p_\theta(z) + (b(z)\nabla_\theta \log p_\theta(z)- b(z)\nabla_\theta \log p_\theta(z))\right] \\
& =\mathbb{E}_z\left[(f(z)- b(z))\nabla_\theta \log p_\theta(z)\right] + \mu_b
\end{align}

We briefly summarize recent stochastic gradient estimators that utilize control variates. We direct the reader to \citet{gu2016} for further detail on these techniques.

\begin{itemize}
\item NVIL \citep{NVIL} uses two baselines: (1) a moving average $\bar{f}$ of $f$ to center the learning signal, and (2) an input-dependent baseline computed by a 1-layer neural network fitted to $f - \bar{f}$ (a control variate for the centered learning signal itself). Finally, variance normalization divides the learning signal by $\max (1,\sigma_f)$, where $\sigma_f^2$ is a moving average of $\text{Var}[f]$. 
\item DARN \citep{DARN} uses $b=f(\bar{z})+f^\prime(\bar{z})(z-\bar{z})$, where the baseline corresponds to the first-order Taylor approximation of $f(z)$ from $f(\bar{z})$. $z$ is chosen to be $\nicefrac{1}{2}$ for Bernoulli variables, which makes the estimator biased for non-quadratic $f$, since it ignores the correction term $\mu_b$ in the estimator expression.
\item MuProp \citep{gu2016} also models the baseline as a first-order Taylor expansion: $b=f(\bar{z})+f^\prime(\bar{z})(z-\bar{z})$ and $\mu_b = f^\prime(\bar{z})\nabla_\theta \mathbb{E}_z\left[z\right]$. To overcome backpropagation through discrete sampling, a mean-field approximation $f_{MF}(\mu_\theta(z))$ is used in place of $f(z)$ to compute the baseline and derive the relevant gradients.
\item VIMCO \citep{VIMCO} is a gradient estimator for multi-sample objectives that uses the mean of other samples $b = \nicefrac{1}{m}\sum_{j\neq i} f(z_j)$ to construct a baseline for each sample $z_i \in z_{1:m}$. We exclude VIMCO from our experiments because we are comparing estimators for single-sample objectives, although Gumbel-Softmax can be easily extended to multi-sample objectives.
\end{itemize} 

\subsection{Semi-Supervised Generative Models}

Semi-supervised learning considers the problem of learning from both labeled data $(x,y) \sim \mathcal{D}_L$ and unlabeled data $x \sim \mathcal{D}_U$, where $x$ are observations (i.e. images) and $y$ are corresponding labels (e.g. semantic class). For semi-supervised classification, \citet{kingma_ssvae} propose a variational autoencoder (VAE) whose latent state is the joint distribution over a Gaussian ``style'' variable $z$ and a categorical ``semantic class'' variable $y$ (Figure \ref{fig:ss_arch}, Appendix). The VAE objective trains a discriminative network $q_\phi(y|x)$, inference network $q_\phi(z|x,y)$, and generative network $p_\theta(x|y,z)$ end-to-end by maximizing a variational lower bound on the log-likelihood of the observation under the generative model. For labeled data, the class $y$ is observed, so inference is only done on $z \sim q(z|x,y)$. The variational lower bound on labeled data is given by:

\begin{equation} \label{eq:var_bound_labeled}
\log p_\theta(x,y) \geq -\mathcal{L}(x,y) = \mathbb{E}_{z \sim q_\phi(z|x,y)}\left[\log p_\theta(x|y,z)] - KL[q(z|x,y)||p_\theta(y)p(z)\right] 
\end{equation}

For unlabeled data, difficulties arise because the categorical distribution is not reparameterizable. \cite{kingma_ssvae} approach this by marginalizing out $y$ over all classes, so that for unlabeled data, inference is still on $q_\phi(z|x,y)$ for each $y$. The lower bound on unlabeled data is:

\begin{align} \label{eq:var_bound_unlabeled}
\log p_\theta(x) \geq -\mathcal{U}(x) & = \mathbb{E}_{z \sim q_\phi(y,z|x)}[\log p_\theta(x|y,z) + \log p_\theta(y) + \log p(z) - q_\phi(y,z|x)]  \\
& = \sum_y q_\phi(y|x) (-\mathcal{L}(x,y) + \mathcal{H}(q_\phi(y|x)))
\end{align}

The full maximization objective is:

\begin{equation} \label{eq:ssvae_objective}
\mathcal{J} = \mathbb{E}_{(x,y) \sim \mathcal{D}_L}\left[-\mathcal{L}(x,y)\right] +  \mathbb{E}_{x \sim \mathcal{D}_U}\left[-\mathcal{U}(x)\right] + \alpha \cdot \mathbb{E}_{(x,y) \sim \mathcal{D}_L}[\log q_\phi(y|x)]
\end{equation}

where $\alpha$ is the scalar trade-off between the generative and discriminative objectives.

One limitation of this approach is that marginalization over all $k$ class values becomes prohibitively expensive for models with a large number of classes. If $D,I,G$ are the computational cost of sampling from $q_\phi(y|x)$, $q_\phi(z|x,y)$, and $p_\theta(x|y,z)$ respectively, then training the unsupervised objective requires $\mathcal{O}(D+k(I+G))$ for each forward/backward step. In contrast, Gumbel-Softmax allows us to backpropagate through $y \sim q_\phi(y|x)$ for single sample gradient estimation, and achieves a cost of $\mathcal{O}(D+I+G)$ per training step. Experimental comparisons in training speed are shown in Figure \ref{fig:ssvae_speed}.

\section{Experimental Results}

In our first set of experiments, we compare Gumbel-Softmax and ST Gumbel-Softmax to other stochastic gradient estimators: Score-Function (SF), DARN, MuProp, Straight-Through (ST), and Slope-Annealed ST. Each estimator is evaluated on two tasks: (1) structured output prediction and (2) variational training of generative models. We use the MNIST dataset with fixed binarization for training and evaluation, which is common practice for evaluating stochastic gradient estimators \citep{SalMurray08,larochelle2011}. 

Learning rates are chosen from $\{3\mathrm{e}{-5},1\mathrm{e}{-5},3\mathrm{e}{-4},1\mathrm{e}{-4},3\mathrm{e}{-3},1\mathrm{e}{-3}\}$; we select the best learning rate for each estimator using the MNIST validation set, and report performance on the test set. Samples drawn from the Gumbel-Softmax distribution are continuous during training, but are discretized to one-hot vectors during evaluation. We also found that variance normalization was necessary to obtain competitive performance for SF, DARN, and MuProp. We used sigmoid activation functions for binary (Bernoulli) neural networks and softmax activations for categorical variables. Models were trained using stochastic gradient descent with momentum $0.9$. 

\subsection{Structured Output Prediction with Stochastic Binary Networks}

The objective of structured output prediction is to predict the lower half of a $28 \times 28$ MNIST digit given the top half of the image ($14 \times 28$). This is a common  benchmark for training stochastic binary networks (SBN) \citep{2014arXiv1406.2989R,gu2016,VIMCO}. The minimization objective for this conditional generative model is an importance-sampled estimate of the likelihood objective, $\mathbb{E}_{h \sim p_\theta(h_i|x_\text{upper})}\left[ \frac{1}{m} \sum_{i=1}^m \log p_\theta(x_\text{lower}|h_i) \right]$, where $m=1$ is used for training and $m=1000$ is used for evaluation.

We trained a SBN with two hidden layers of 200 units each. This corresponds to either 200 Bernoulli variables (denoted as $392$-$200$-$200$-$392$) or 20 categorical variables (each with 10 classes) with binarized activations (denoted as $392$-$(20 \times 10)$-$(20 \times 10)$-$392$).

\begin{figure}[h]
  \centering
	\subfigure[]{\includegraphics[width=.49\textwidth]{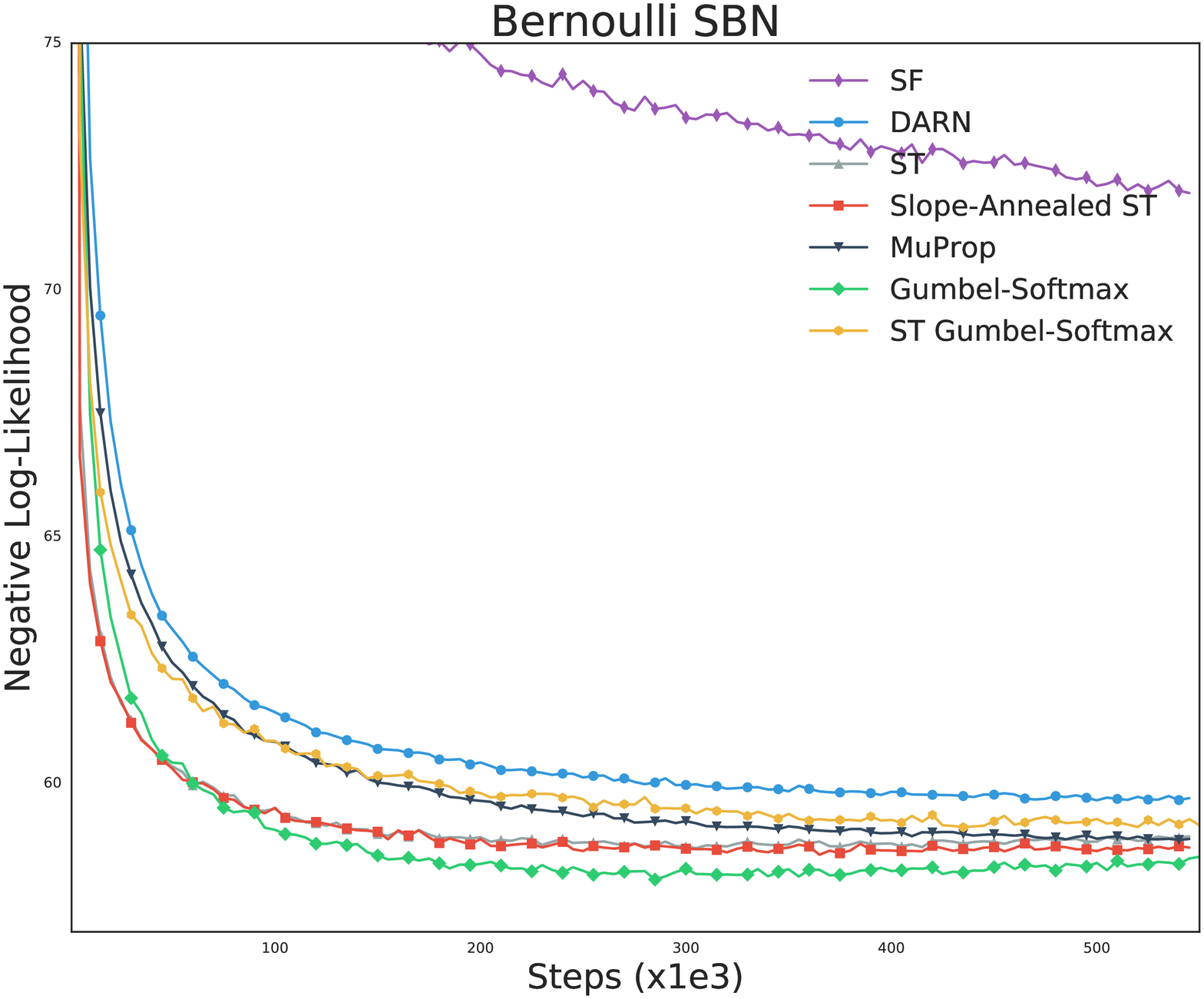}}
 	\subfigure[]{\includegraphics[width=.49\textwidth]{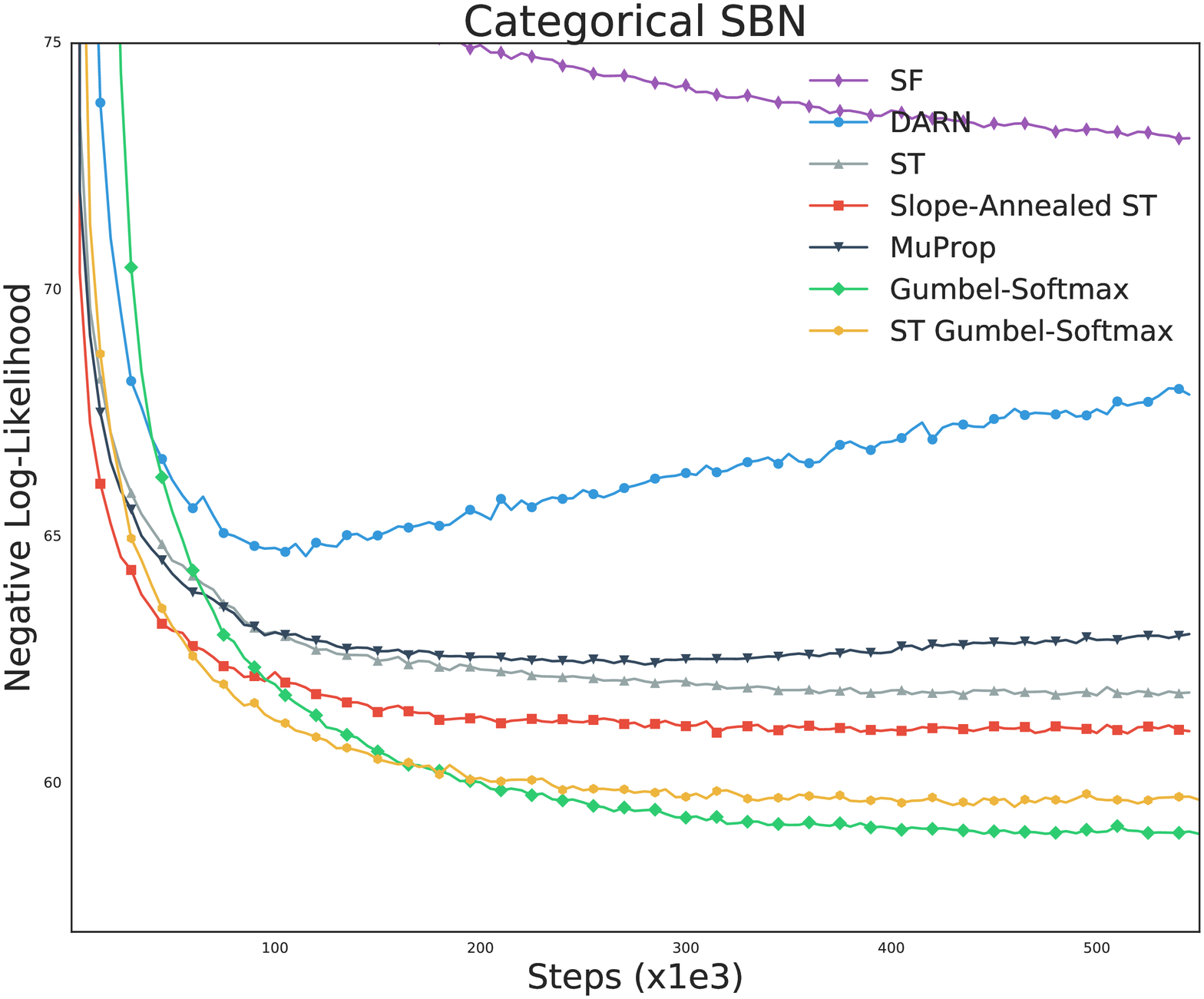}}
  \caption{Test loss (negative log-likelihood) on the structured output prediction task with binarized MNIST using a stochastic binary network with (a) Bernoulli latent variables ($392$-$200$-$200$-$392$) and (b) categorical latent variables ($392$-$(20 \times 10)$-$(20 \times 10)$-$392$). }
  \label{fig:sbn_curves}
\end{figure}

As shown in Figure \ref{fig:sbn_curves}, ST Gumbel-Softmax is on par with the other estimators for Bernoulli variables and outperforms on categorical variables. Meanwhile, Gumbel-Softmax outperforms other estimators on both Bernoulli and Categorical variables. We found that it was not necessary to anneal the softmax temperature for this task, and used a fixed $\tau=1$.

\subsection{Generative Modeling with Variational Autoencoders}

We train variational autoencoders \citep{kingma_vae}, where the objective is to learn a generative model of binary MNIST images. In our experiments, we modeled the latent variable as a single hidden layer with 200 Bernoulli variables or 20 categorical variables ($20 \times 10$). We use a learned categorical prior rather than a Gumbel-Softmax prior in the training objective. Thus, the minimization objective during training is no longer a variational bound if the samples are not discrete. In practice, we find that optimizing this objective in combination with temperature annealing still minimizes actual variational bounds on validation and test sets. Like the structured output prediction task, we use a multi-sample bound for evaluation with $m=1000$.

The temperature is annealed using the schedule $\tau = \text{max}(0.5,\text{exp}(-rt))$ of the global training step $t$, where $\tau$ is updated every $N$ steps. $N \in \{500,1000\}$ and $r \in \{1\mathrm{e}{-5},1\mathrm{e}{-4}\}$ are hyperparameters for which we select the best-performing estimator on the validation set and report test performance. 

\begin{figure}[h] 
  \centering
	\subfigure[]{\includegraphics[width=.49\textwidth]{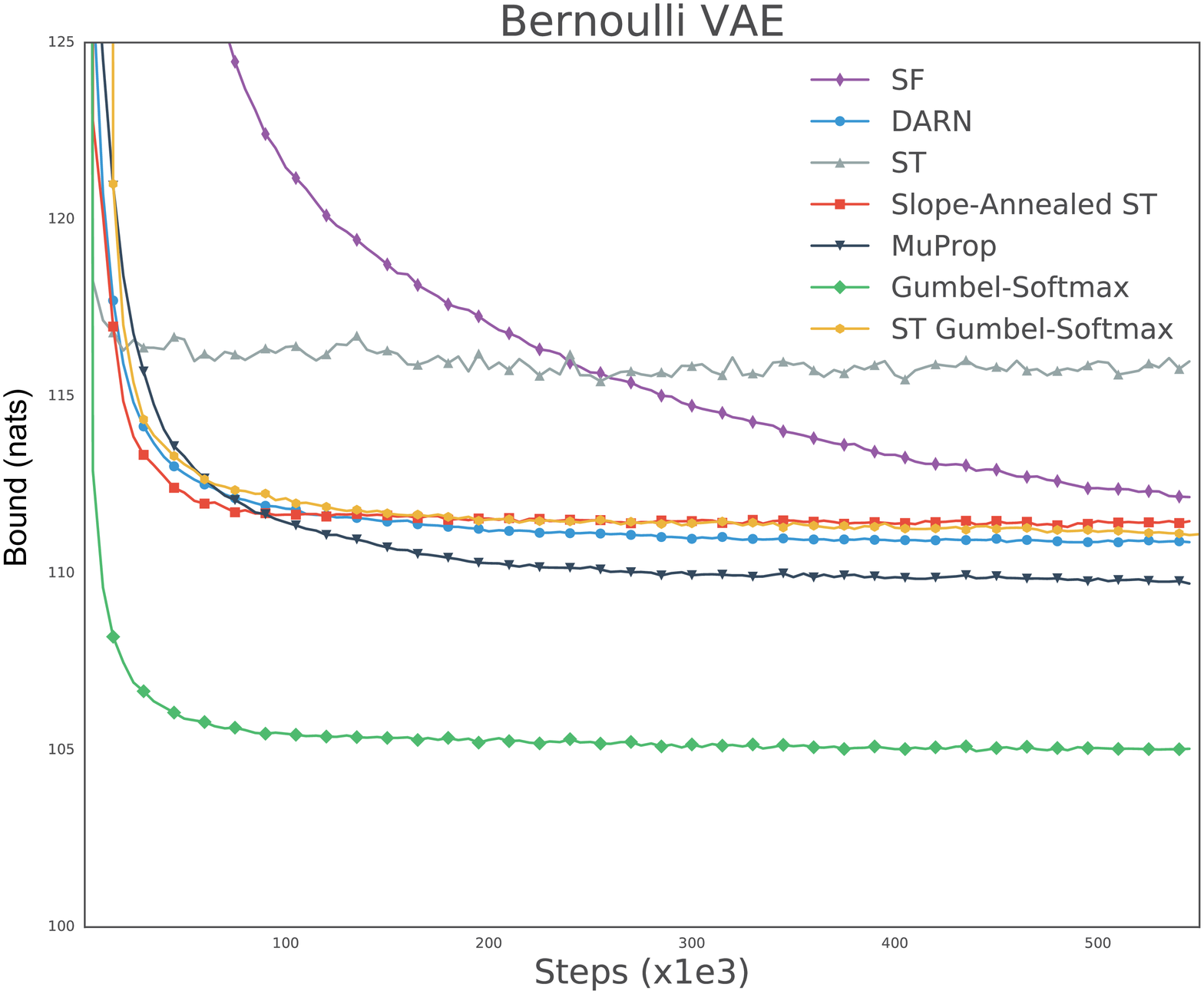}}
 	\subfigure[]{\includegraphics[width=.49\textwidth]{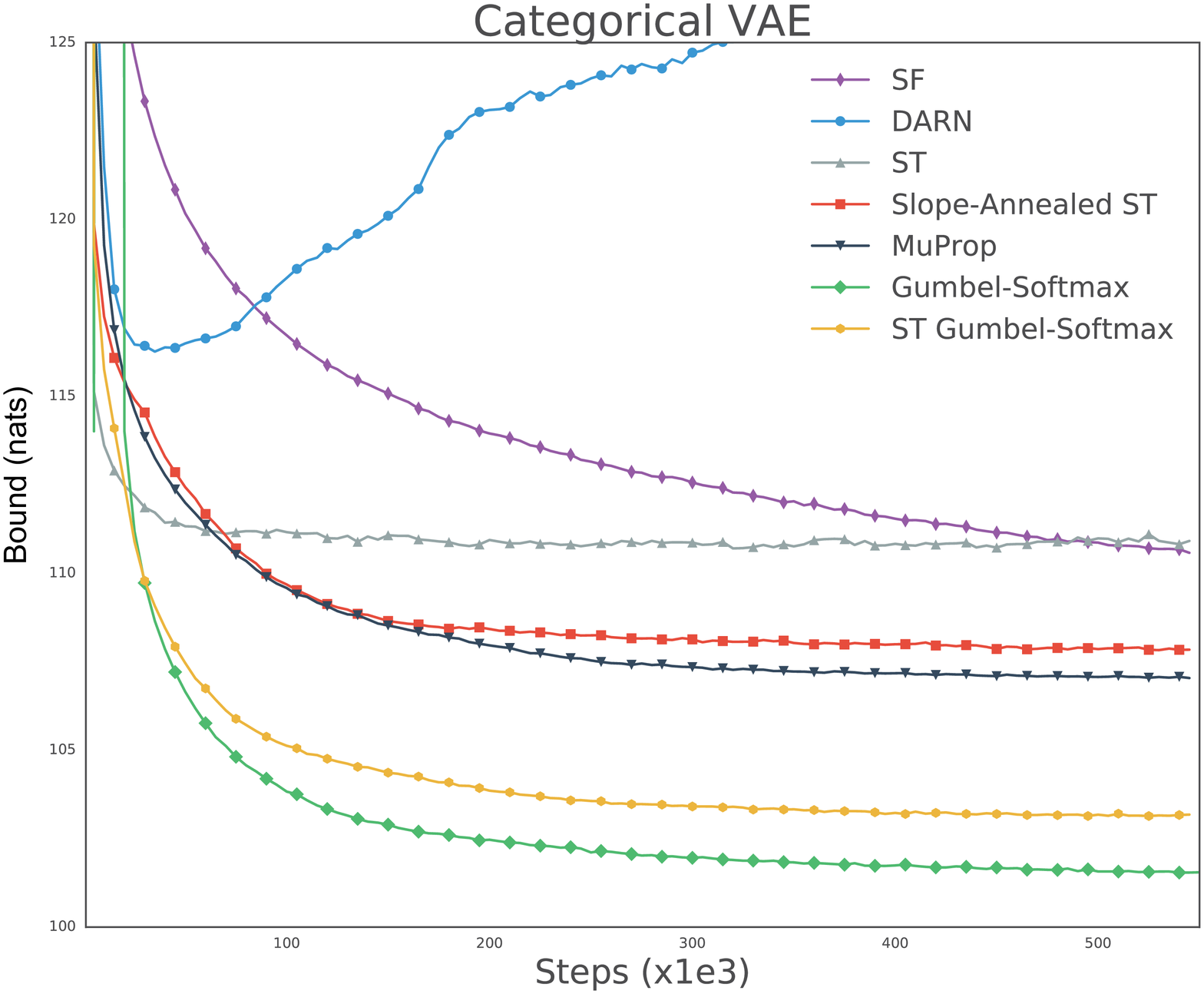}}
  \caption{Test loss (negative variational lower bound) on binarized MNIST VAE with (a) Bernoulli latent variables ($784-200-784$) and (b) categorical latent variables ($784-(20\times 10)-200$).}
  \label{fig:vae_curves}
\end{figure}

As shown in Figure \ref{fig:vae_curves}, ST Gumbel-Softmax outperforms other estimators for Categorical variables, and Gumbel-Softmax drastically outperforms other estimators in both Bernoulli and Categorical variables.

\begin{table}[ht]
\caption{The Gumbel-Softmax estimator outperforms other estimators on Bernoulli and Categorical latent variables. For the structured output prediction (SBN) task, numbers correspond to negative log-likelihoods (nats) of input images (lower is better). For the VAE task, numbers correspond to negative variational lower bounds (nats) on the log-likelihood (lower is better).}
\label{tab:results}
\begin{center}
 \begin{tabular}{c|c c c c c c c}
  				& SF & DARN & MuProp & ST & Annealed ST & Gumbel-S. & ST Gumbel-S. \\
 \hline
SBN (Bern.)	 & 72.0	& 59.7	& 58.9	& 58.9	& 58.7	& \textbf{58.5} & 59.3 \\
SBN (Cat.) & 73.1	 & 67.9 & 63.0 & 61.8 & 61.1	& \textbf{59.0}	& 59.7 \\
VAE (Bern.)	& 112.2	& 110.9	& 109.7	& 116.0	& 111.5	& \textbf{105.0}	& 111.5 \\
VAE (Cat.) &	110.6 & 128.8 & 107.0 & 110.9 & 107.8 & \textbf{101.5} & 107.8
  \end{tabular}
\end{center}
\end{table}

\subsection{Generative Semi-Supervised Classification}
\label{section:ssclass}

We apply the Gumbel-Softmax estimator to semi-supervised classification on the binary MNIST dataset. We compare the original marginalization-based inference approach \citep{kingma_ssvae} to single-sample inference with Gumbel-Softmax and ST Gumbel-Softmax. 

We trained on a dataset consisting of 100 labeled examples (distributed evenly among each of the 10 classes) and 50,000 unlabeled examples, with dynamic binarization of the unlabeled examples for each minibatch. The discriminative model $q_\phi(y|x)$ and inference model $q_\phi(z|x,y)$ are each implemented as 3-layer convolutional neural networks with ReLU activation functions. The generative model $p_\theta(x|y,z)$ is a 4-layer convolutional-transpose network with ReLU activations. Experimental details are provided in Appendix \ref{appendix:conv_arch}.

Estimators were trained and evaluated against several values of $\alpha = \{0.1,0.2,0.3,0.8,1.0\}$ and the best unlabeled classification results for test sets were selected for each estimator and reported in Table \ref{tab:ssvae_results_mnist}. We used an annealing schedule of $\tau = \text{max}(0.5, \text{exp}(-3\mathrm{e}{-5}\cdot t))$, updated every 2000 steps.

\begin{figure}[h]  
  \centering
  \subfigure[]{\includegraphics[width=.45\textwidth]{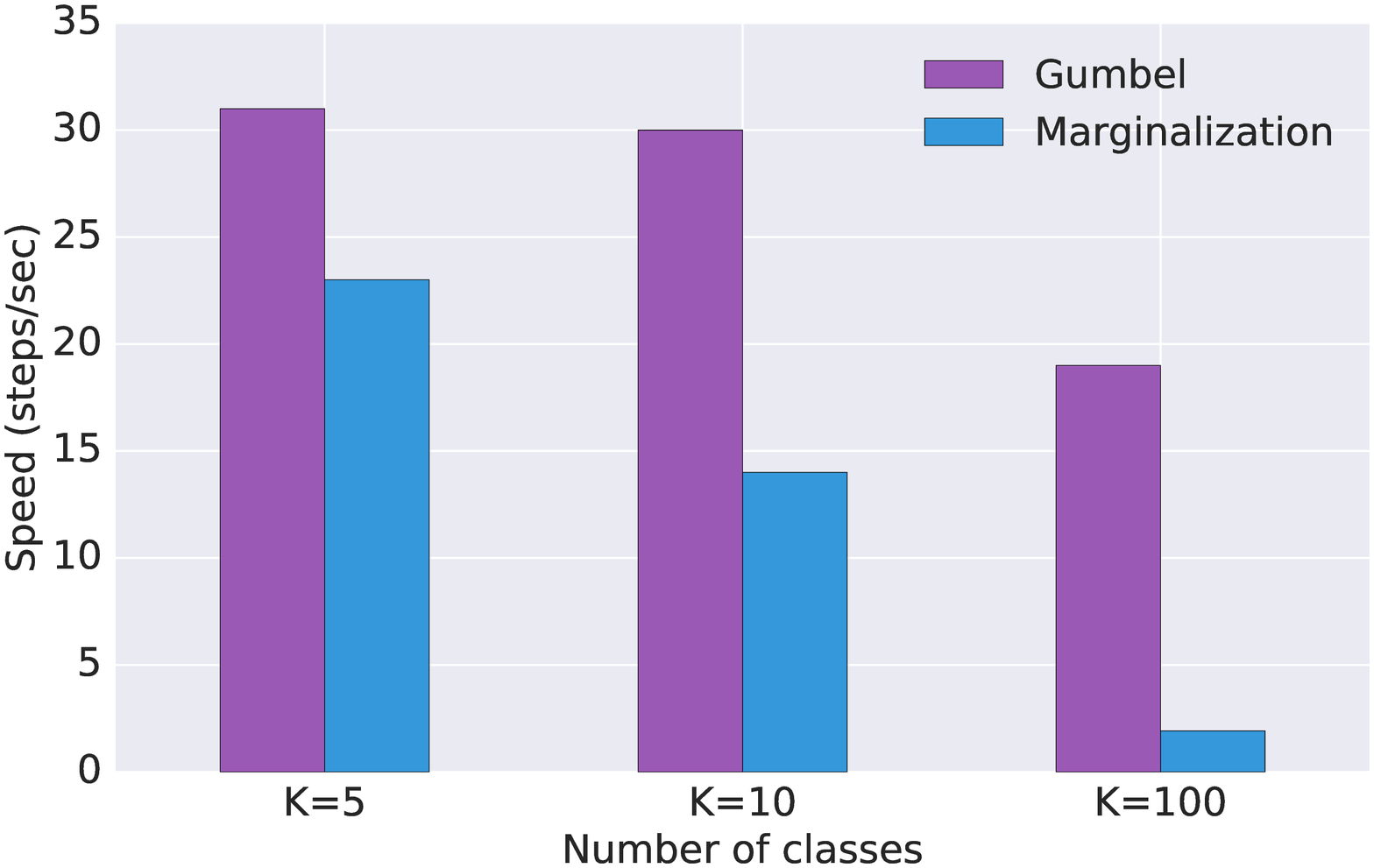}}
  \subfigure[]{\includegraphics[width=.45\textwidth]{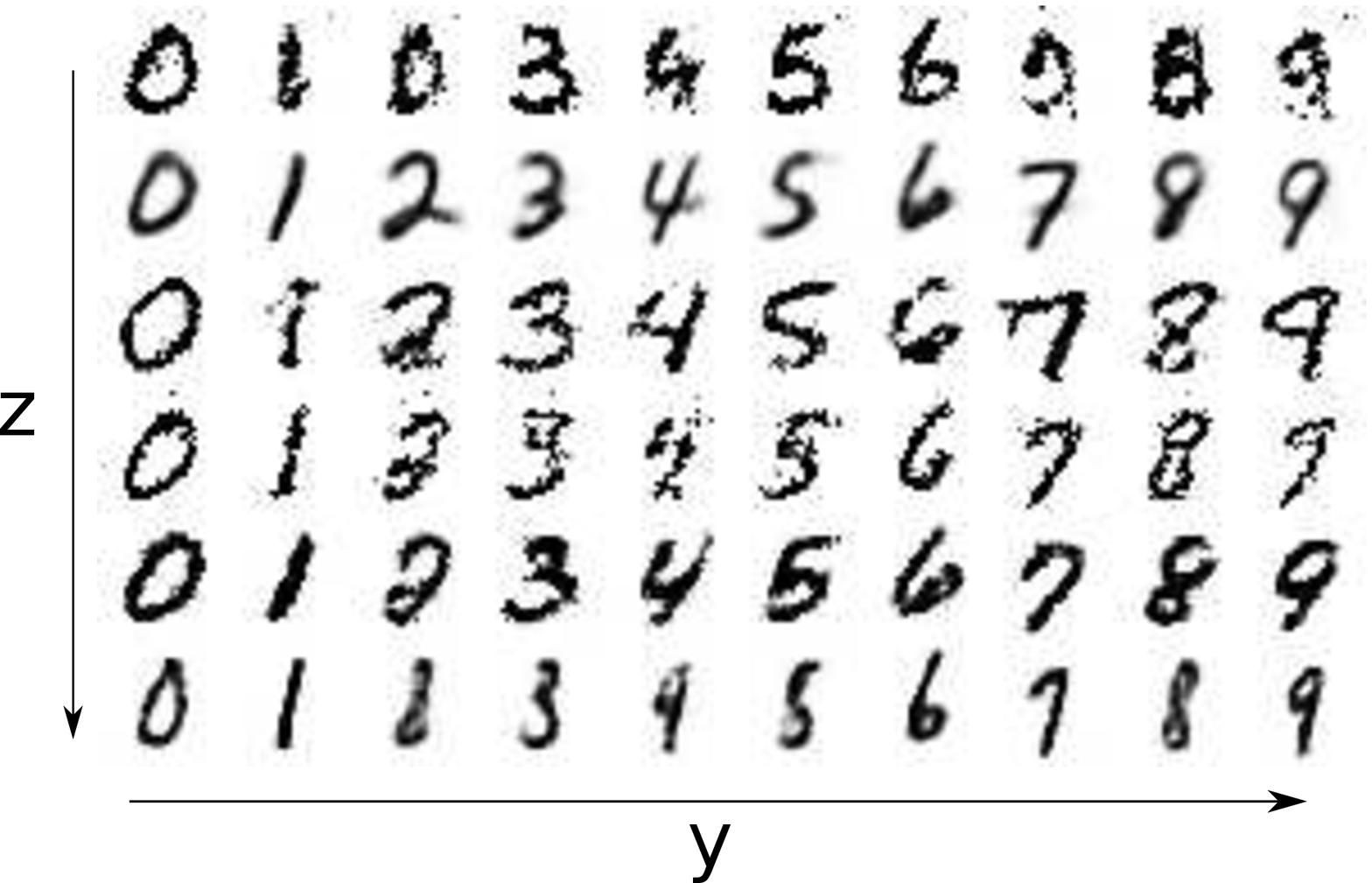}}
  \caption{Gumbel-Softmax allows us to backpropagate through samples from the posterior $q_\phi(y|x)$, providing a scalable method for semi-supervised learning for tasks with a large number of classes. (a) Comparison of training speed (steps/sec) between Gumbel-Softmax and marginalization \citep{kingma_ssvae} on a semi-supervised VAE. Evaluations were performed on a GTX Titan X\textsuperscript{\textregistered} GPU. (b) Visualization of MNIST analogies generated by varying style variable $z$ across each row and class variable $y$ across each column.}
   \label{fig:ssvae_speed}
\end{figure}

In \citet{kingma_ssvae}, inference over the latent state is done by marginalizing out $y$ and using the reparameterization trick for sampling from $q_\phi(z|x,y)$. However, this approach has a computational cost that scales linearly with the number of classes. Gumbel-Softmax allows us to backpropagate directly through single samples from the joint $q_\phi(y,z|x)$, achieving drastic speedups in training without compromising generative or classification performance. (Table \ref{tab:ssvae_results_mnist}, Figure \ref{fig:ssvae_speed}).

\begin{table}[ht]
\caption{Marginalizing over $y$ and single-sample variational inference perform equally well when applied to image classification on the binarized MNIST dataset \citep{larochelle2011}. We report variational lower bounds and image classification accuracy for unlabeled data in the test set.}
\label{tab:ssvae_results_mnist}
\begin{center}
\begin{tabular}{l|l|l}
					 & ELBO & Accuracy \\
   \hline
Marginalization        & -106.8 & 92.6\% \\
Gumbel                & -109.6 & 92.4\% \\
ST Gumbel-Softmax    & -110.7 & 93.6\%
\end{tabular}
\end{center}
\end{table}

In Figure \ref{fig:ssvae_speed}, we show how Gumbel-Softmax versus marginalization scales with the number of categorical classes. For these experiments, we use MNIST images with randomly generated labels. Training the model with the Gumbel-Softmax estimator is $2\times$ as fast for $10$ classes and $9.9\times$ as fast for $100$ classes.

\section{Discussion}
The primary contribution of this work is the reparameterizable Gumbel-Softmax distribution, whose corresponding estimator affords low-variance path derivative gradients for the categorical distribution. We show that Gumbel-Softmax and Straight-Through Gumbel-Softmax are effective on structured output prediction and variational autoencoder tasks, outperforming existing stochastic gradient estimators for both Bernoulli and categorical latent variables. Finally, Gumbel-Softmax enables dramatic speedups in inference over discrete latent variables.

\subsubsection*{Acknowledgments}
We sincerely thank Luke Vilnis, Vincent Vanhoucke, Luke Metz, David Ha, Laurent Dinh, George Tucker, and Subhaneil Lahiri for helpful discussions and feedback.

\bibliography{iclr2017_conference}
\bibliographystyle{iclr2017_conference}

\appendix

\section{Semi-Supervised Classification Model}
\label{appendix:conv_arch}

\begin{figure}[ht]
  \centering
\includegraphics[width=.7\textwidth]{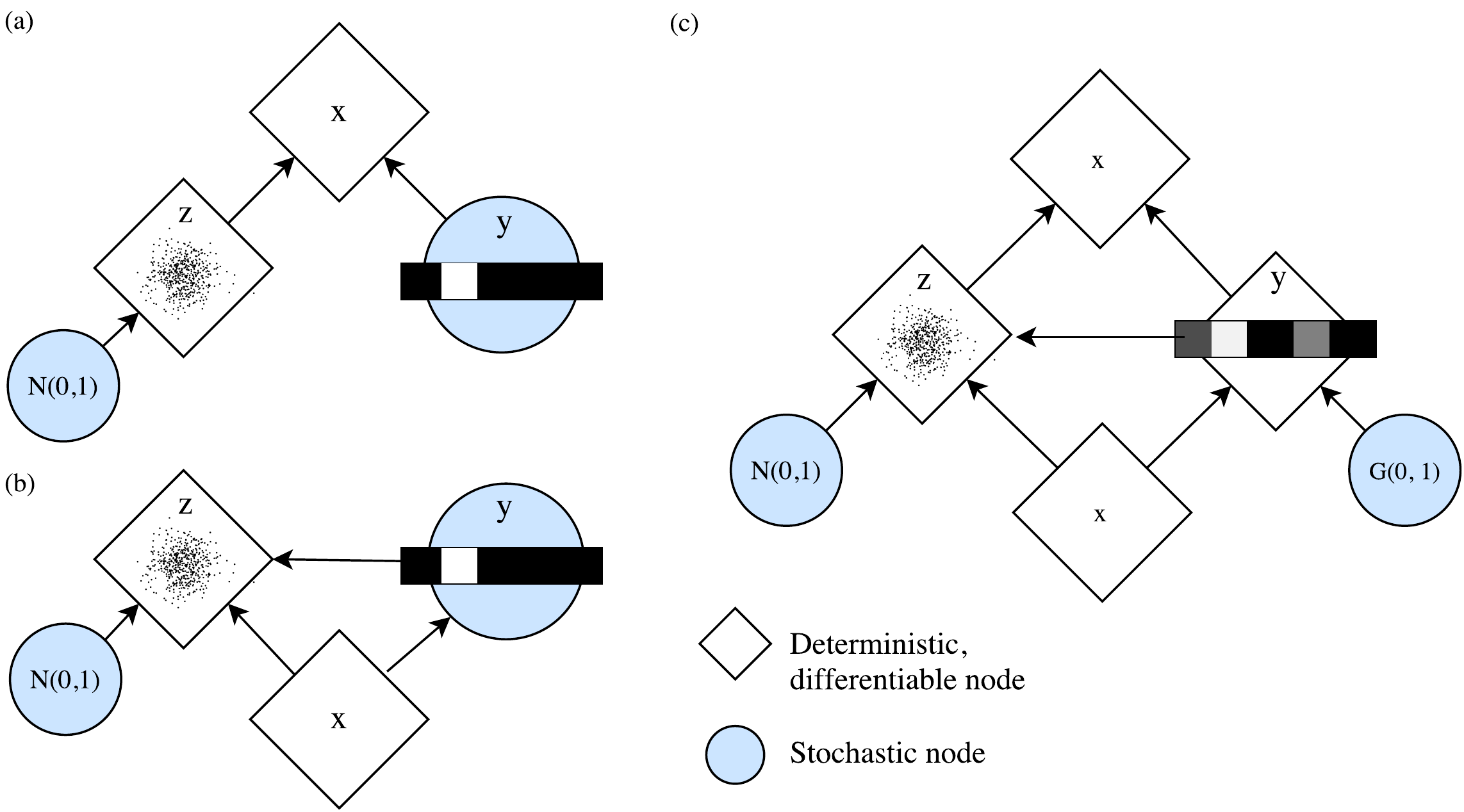}
  \caption{Semi-supervised generative model proposed by \citet{kingma_ssvae}. (a) Generative model $p_\theta(x|y,z)$ synthesizes images from latent Gaussian ``style'' variable $z$ and categorical class variable $y$. (b) Inference model $q_\phi(y,z|x)$ samples latent state $y,z$ given $x$. Gaussian $z$ can be differentiated with respect to its parameters because it is reparameterizable. In previous work, when $y$ is not observed, training the VAE objective requires marginalizing over all values of $y$. (c) Gumbel-Softmax reparameterizes $y$ so that backpropagation is also possible through $y$ without encountering stochastic nodes.} 
 \label{fig:ss_arch}
\end{figure}

\begin{figure}[ht] 
  \centering
\includegraphics[width=.7\textwidth]{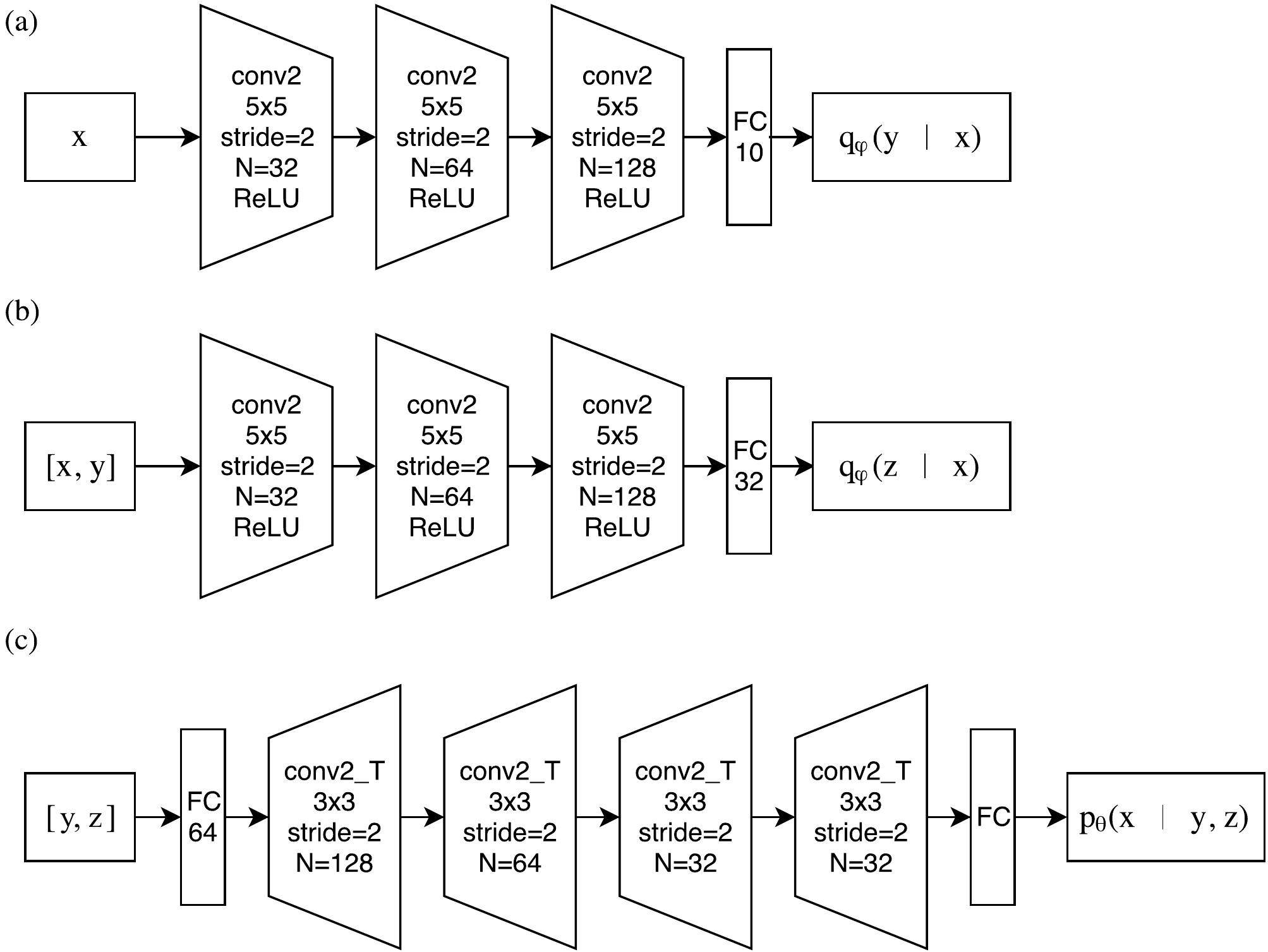}
  \caption{Network architecture for (a) classification $q_\phi(y|x)$ (b) inference $q_\phi(z|x,y)$, and (c) generative $p_\theta(x|y,z)$ models. The output of these networks parameterize Categorical, Gaussian, and Bernoulli distributions which we sample from.}
  \label{fig:conv_arch}
\end{figure}

Figures \ref{fig:ss_arch} and \ref{fig:conv_arch} describe the architecture used in our experiments for semi-supervised classification (Section \ref{section:ssclass}).

\section{Deriving the density of the Gumbel-Softmax distribution}
\label{appendix:gumbel_derivation}

 Here we derive the probability density function of the Gumbel-Softmax distribution with probabilities $\pi_1, ..., \pi_k$ and temperature $\tau$. We first define the logits $x_i=\log \pi_i$, and Gumbel samples $g_1, ..., g_k$, where $g_i \sim \text{Gumbel}(0, 1)$. A sample from the Gumbel-Softmax can then be computed as:
 \begin{equation}
y_i = \frac{\exp\left((x_i + g_i) / \tau\right)  }{\sum_{j=1}^k \exp\left((x_j + g_j)/\tau \right)}\qquad \text{for } i =1,..., k
\end{equation}

\subsection{Centered Gumbel density}
The mapping from the Gumbel samples $g$ to the Gumbel-Softmax sample $y$ is not invertible as the normalization of the softmax operation removes one degree of freedom. To compensate for this, we define an equivalent sampling process that subtracts off the last element, $(x_k + g_k) / \tau$ before the softmax:
\begin{equation}
y_i = \frac{\exp\left((x_i + g_i - (x_k + g_k)) / \tau\right)  }{\sum_{j=1}^k \exp\left((x_j + g_j - (x_k + g_k))/\tau \right)}\qquad \text{for } i =1,..., k
\end{equation}

To derive the density of this equivalent sampling process, we first derive the density for the "centered" multivariate Gumbel density corresponding to:
\begin{equation}
u_i = x_i + g_i - (x_k + g_k) \qquad \text{for } i = 1, ..., k-1
\end{equation}
where $g_i \sim \text{Gumbel}(0, 1)$. Note the probability density of a Gumbel distribution with scale parameter $\beta=1$ and mean $\mu$ at $z$ is: $f(z, \mu) = e^{\mu-z-e^{\mu-z}}$. We can now compute the density of this distribution by marginalizing out the last Gumbel sample, $g_k$:
\begin{align*}
p(u_1, ..., u_{k-1}) &= \int_{-\infty}^\infty d g_k\, p(u_1, ..., u_{k} | g_k) p(g_k)\\
&=\int_{-\infty}^\infty d g_k\, p(g_k) \prod_{i=1}^{k-1} p(u_i | g_k)\\
&= \int_{-\infty}^\infty d g_k\, f(g_k, 0) \prod_{i=1}^{k-1} f(x_k + g_k, x_i - u_i )\\
&= \int_{-\infty}^\infty d g_k\, e^{ - g_k - e^{ - g_k}} \prod_{i=1}^{k-1} e^{x_i - u_i - x_k-g_k - e^{x_i - u_i - x_k-g_k}}
\end{align*}
We perform a change of variables with $v=e^{-g_k}$, so $dv=-e^{-g_k}dg_k$ and $dg_k = -dv\, e^{g_k}=dv/v$, and define $u_k=0$ to simplify notation:
\begin{align}\label{eq:gumbel_centered}
p(u_1, ..., u_{k, -1})&= \delta(u_k=0)\int_{0}^\infty d v\, \frac{1}{v} ve^{x_k - v} \prod_{i=1}^{k-1} ve^{x_i - u_i-x_k  - ve^{x_i - u_i-x_k}}\\
&= \exp\left(x_k+\sum_{i=1}^{k-1}(x_i - u_i)\right)\left(e^{x_k} + \sum_{i=1}^{k-1} \left(e^{x_i - u_i}\right)\right)^{-k} \Gamma(k)\\
&= \Gamma(k)\exp\left(\sum_{i=1}^{k}(x_i - u_i)\right)\left(\sum_{i=1}^{k} \left(e^{x_i - u_i}\right)\right)^{-k}\\
&= \Gamma(k)\left(\prod_{i=1}^k\exp\left(x_i - u_i\right)\right)\left(\sum_{i=1}^{k} \exp\left(x_i - u_i\right)\right)^{-k}
\end{align}

\subsection{Transforming to a Gumbel-Softmax}
Given samples $u_1, ..., u_{k, -1}$ from the centered Gumbel distribution, we can apply a deterministic transformation $h$ to yield the first $k-1$ coordinates of the sample from the Gumbel-Softmax:
\begin{equation}
y_{1:k-1} = h(u_{1:k-1}), \qquad h_i(u_{1:k-1}) = \frac{\exp(u_i / \tau)}{1 + \sum_{j=1}^{k-1} \exp(u_j / \tau)} \forall i=1,\dots,{k-1}
\end{equation}
Note that the final coordinate probability $y_k$ is fixed given the first $k-1$, as $\sum_{i=1}^k y_i=1$:
\begin{equation}
y_k = \left(1 + {\sum_{j=1}^{k-1} \exp(u_j / \tau)}\right)^{-1} = 1 - \sum_{j=1}^{k-1} y_j
\end{equation}

We can thus compute the probability of a sample from the Gumbel-Softmax using the change of variables formula on only the first $k-1$ variables:
\begin{equation}\label{eq:change_of_var}
p(y_{1:k}) = p\left(h^{-1}(y_{1:k-1}) \right)\text{det}\left(\frac{\partial h^{-1}(y_{1:k-1}) }{\partial y_{1:k-1}}\right)
\end{equation}
Thus we need to compute two more pieces: the inverse of $h$ and its Jacobian determinant.
The inverse of $h$ is:
\begin{align}\label{eq:h_inv}
h^{-1}(y_{1:k-1})= \tau\times \left(\log y_i - \log\left(1 - \sum_{j=1}^{k-1}y_j\right)\right) = \tau \times \left(\log y_i - \log y_k \right)
\end{align}
with Jacobian
\begin{equation}
\frac{\partial h^{-1}(y_{1:k-1}) }{\partial y_{1:k-1}} = \tau \times \left(\text{diag}\left(\frac{1}{y_{1:k-1}}\right)  + \frac{1}{y_k}\right)=
\begin{bmatrix}
\frac{1}{y_1} + \frac{1}{y_k} & \frac{1}{y_k}&\dots & \frac{1}{y_k}\\
\frac{1}{y_k} & \frac{1}{y_2} + \frac{1}{y_k} &\dots &  \frac{1}{y_k}\\
\vdots & \vdots & \ddots & \vdots\\
\frac{1}{y_k} & \frac{1}{y_k} &\dots & \frac{1}{y_{k-1}} + \frac{1}{y_k}
\end{bmatrix}
\end{equation}
Next, we compute the determinant of the Jacobian:
\begin{align}
\text{det}\left(\frac{\partial h^{-1}(y_{1:k-1}) }{\partial y_{1:k-1}} \right) &= \tau^{k-1}\text{det}\left(\left(I + \frac{1}{y_k}ee^T\text{diag}\left(y_{1:k-1}\right)\right) \left(\text{diag}\left(\frac{1}{y_{1:k-1}}\right)\right)\right)\\
&= \tau^{k-1}\left(1 + \frac{1-y_k}{y_k}\right)\prod_{j=1}^{k-1} y_j^{-1}\\
&= \tau^{k-1}\prod_{j=1}^{k} y_j^{-1}
\label{eq:h_jac}
\end{align}
where $e$ is a $k-1$ dimensional vector of ones, and we've used the identities: $\text{det}(AB)=\text{det}(A)\text{det}(B)$, $\text{det}(\text{diag}(x))=\prod_i x_i$, and $\text{det}(I+uv^T)=1+u^Tv$.

We can then plug into the change of variables formula (Eq. \ref{eq:change_of_var}) using the density of the centered Gumbel (Eq.\ref{eq:gumbel_centered}), the inverse of $h$ (Eq. \ref{eq:h_inv}) and its Jacobian determinant (Eq. \ref{eq:h_jac}):
\begin{align}
p(y_1, .., y_k) &= \Gamma(k)\left(\prod_{i=1}^k\exp\left(x_i\right)\frac{y_k^\tau}{y_i^\tau}\right)\left(\sum_{i=1}^{k} \exp\left(x_i\right)\frac{y_k^\tau}{y_i^\tau}\right)^{-k} \tau^{k-1}\prod_{i=1}^{k} y_i^{-1}\\
&= \Gamma(k)\tau^{k-1}\left(\sum_{i=1}^{k} \exp\left(x_i\right)/{y_i^\tau}\right)^{-k} \prod_{i=1}^k\left(\exp\left(x_i\right)/y_i^{\tau+1}\right)
\end{align}
\end{document}